# DESIGN AND IMPLEMENTATION OF PROSTHETIC ARM USING GEAR MOTOR CONTROL TECHNIQUE WITH APPROPRIATE TESTING

[1]BISWARUP NEOGI, [2] SOUMYAJIT MUKHERJEE, [3] SOUMYA GHOSAL,
[4] DR. ACHINTYA DAS, [5] D.N.TIBAREWALA.

[3]Sr. Lecturer, Electronics & Communication Department, Durgapur Institute of Advanced Technology & Management, Durgapur
[2]CSE Department, Saroj Mohan Institute of Technology, West Bengal, India.
[1]IT Department, RCC Institute of Information Technology, Kolkata, India.
[4]Professor & HOD, Electronics & Communication Department, Kalyani Government Engineering College.
[5]Director, School of Bio Science & Engineering, Jadavpur University, Kolkata,W.B, India.

*biswarupneogi@gmail.com, soumyajitmukherjee.cs@gmail.com, soumyaghosal@ieee.org, achintya_das123@yahoo.com, biomed_ju@yahoo.com*

*ABSTRACT :*
*Any part of the human body replication procedure commences the prosthetic control science. This paper highlights the hardware design technique of a prosthetic arm with implementation of gear motor control aspect. The prosthetic control arm movement has been demonstrated in this paper applying processor programming and with the successful testing of the designed prosthetic model. The architectural design of the prosthetic arm here has been replaced by lighter material instead of heavy metal, as well as the traditional EMG (electro myographic) signal has been replaced by the muscle strain.*

*Keywords: Prosthetic arm, PIC, EMG Signal, Muscle strain.*

## 1. INTRODUCTION

The earlier research works developed so far on the prosthetic limb failed to consider the cost effectiveness of the product. Also those designs were uncongenial for movement due to the use of heavier metal in constructing the arm instead of lighter ones. The man machine interface process is easier to theorize but its practical implementation tends to be much tougher. This work, directly related to the handicapped patients makes it a potential aid for the society. In essence, this device makes the prosthetic limb to mimic a real limb, restoring the associated functionalities and efficacies of natural arm movement [3][6]. We succeeded in devising a prosthetic arm which can be easily controlled by human limb, and used a PIC microcontroller system to control the prosthetic arm associated with a sensor to get information from human limb. This input signal can easily be controlled by the muscle movement activating the action of either griping or stretching. This was followed by employing a metallic human hand type arm aiming to implement automation technique. A successful implementation of the project relating the control of prosthetic arm through microcontroller programming would be of much help to the mankind who lost their arms.

## 2. TECHNOLOGY SURVEY BEHIND PROSTHETIC ARM

The main technical concept, behind this starts from taking the signal from the movement of our limb and with the help of microcontroller program of the signal; we are able to create the movement of the artificial arm. New technique that capitalizes on the movement of remaining nerves allows amputees to intuitively control their prosthetic limb, providing them with a much better level of control than traditional prosthetics. The rerouted nerves growing in the muscle, amplified the messages once sent to muscles in the arm.

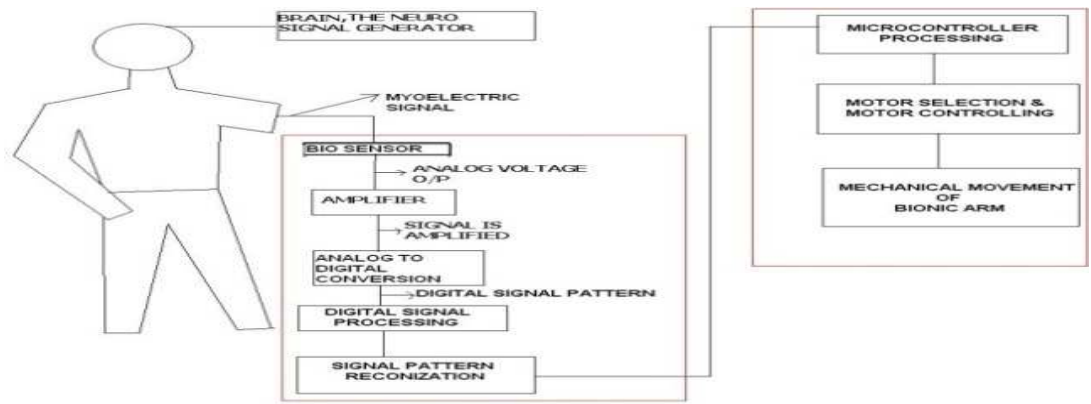

**Figure. 1A. Block Diagram of EMG Controlled Arm**

These signals are read by sensors on the prosthetic limb and translated into movement. It is highly encouraging to observe that even after an amputation; the same intention to move the limb can be harnessed to control a prosthetic limb in much the same way as the limb was previously controlled. Most artificial arms, controlled by remaining muscles near the amputated limb but the devices can be frustrating and slow.

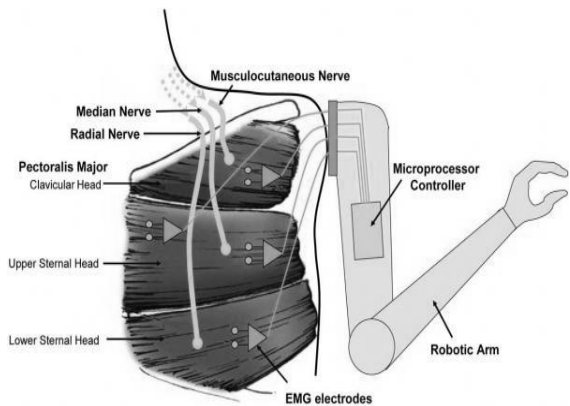

**Figure. 1B Schematics Diagram of Block Prosthesis**

frustrating and slow. The user must consciously contract those muscles to trigger a movement, resulting single movement performed at a time. More intuitive method can be initiated for controlling prosthetics that capitalize on remaining nerves, which still carry neural signals meant for the lost limb. Fig. 1A represents the basic block diagram of prosthetic arm. The biosensor is a special type of sensor which captures the EMG signal from a particular muscle of human body. Generally the EMG signal is in micro-volt white noise signal which can be amplified through a perfect high CMRR amplifier. The amplifier output is converted by ADC with appropriate processing of EMG signals, interpreted by the pattern recognition technique. The microcontroller is capable to accept the digitized form of EMG command of arm movement. The processor produces the movement of motor with the formal action of human muscle. The movement of the motor controls the gear for high torque generation at the same time maintains the linearity for prosthetic arm movement. Fig. 1B represents the schematic representation of the system [4].

**3. LIMITATION OF MYOELECTRIC SIGNAL SENSING AND INTERPRETATION**

The myoelectric signal is generated by the brain to activate the muscles of human body. In fact the nature of eletro-myograpic (EMG) signal is very much like to that of a wide Gaussian noise. It is very difficult to sense and interpret the myoelectric from muscle. However many scientists worked in this field and generated EMG controlled artificial arm. Among them the significant EMG signal controlled prosthetic arm was generated by Eugene J. Moore, Robert M. De Marco, Richard A. Foulds and Tara L. Alvarez in Department of Biomedical Engineering, New Jersey Institute of Technology, Newark, New Jersey [5]. In this context it may be mentioned that EMG controlled arm is an intelligent arm which therefore can be controlled by the incoming brain message property. However this type of control includes high expenses due to the construction of adjoining sensor and amplifier. Mention may also be made to the fact, in tropical country like India due to excess secretion perspiration through the pours of skin gives rise to extra noisy bio-potential creating problem to the EMG signal sensor. That is why this type of EMG signal controlled artificial arm is not suitable for our country. Owing to all these reasons stated above we intend to develop a different type of controlled arm which is acting with the muscle pressure. This type of design is activated by the pressure developed in the muscle which will stay unaffected by the formation of the sweat on the skin. In this System the two tact switches are used as a sensor which are comparatively much cheaper. The tact switches produce the signal which is in digitized form, so we do not use the ADC keeping the price even low. Clinically we have recorded the EMG signals of four different positions of the arm muscle S1,S2,S3,S4 represented in Fig.2) for stress condition vide Fig. 3B and without stress condition vide fig 3A. The EMG signals are recorded for without strain condition of arm represents in Fig. 3A

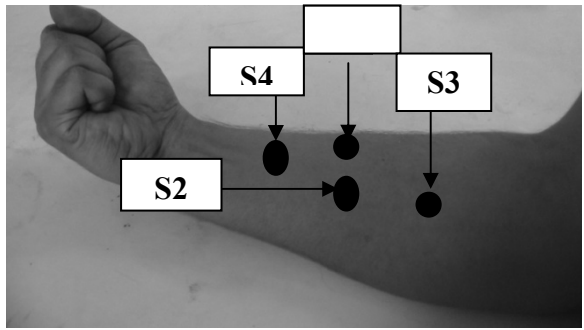

Fig2. The Position of EMG Signal Sensors

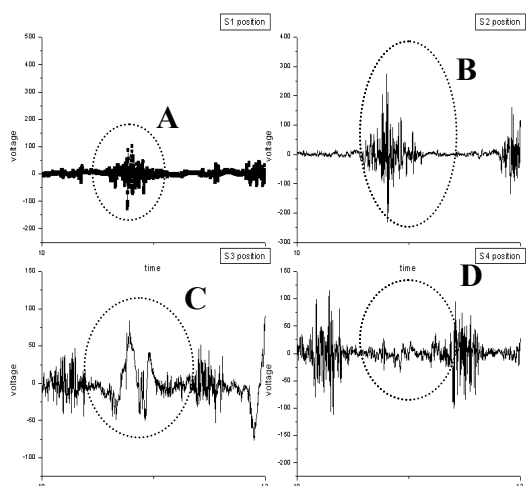

Figure 3A. The EMG Signal Pattern of S1, S2, S3 and S4 Position in without Stress condition of arm

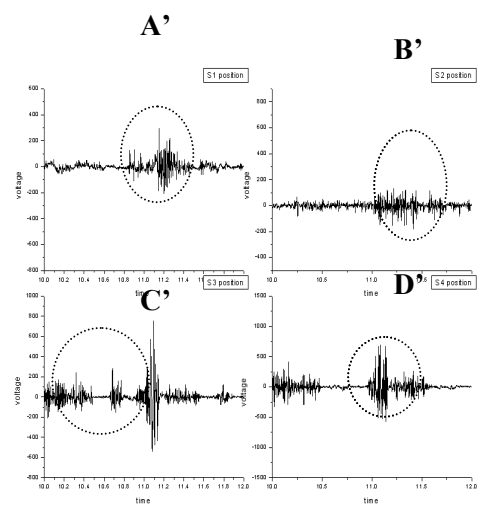

### Figure 3B. The EMG Signal Pattern of S1, S2, S3and S4 Position in with Stress condition of arm

After applying the strain in the muscle with the wrist activity of arm the EMG signals recorded are represented in Fig 3B. From 30029 time range of data we are representing only the data from 10 to 12 time units. Here it is to be remembered that the voltage of EMG signals are in micro volt range. From Fig. 2A S1 it is seen that the recorded EMG signal without stressed condition is quite appreciable at the position A where as the in the Fig. 3B S1 the corresponding position A' the signal is quite inappreciable. Also in the stressed condition the mean value of the potential at position B' is almost zero whereas the corresponding position B (without stressed condition) is quite. Moreover, in Fig. 2A S3 the position C represents the higher potential than the position C' in Fig 3B S3. In comparison with the previously discussed figures (Fig. 3A S1 and Fig. 3B S1, Fig. 3A S2 and Fig. 3B S2, Fig. 3A S3 and Fig. 3B S3 ), the Fig. 3A S4 and 3B S4 show different characteristics, i.e. both the D (Fig. 3A S1) and the D' (Fig. 3B S2) do not exhibit any differences. These signals are very difficult to interpret; that is why we are deserting aside these type of EMG signals and opting to myo-control to muscle pressure [4][8].

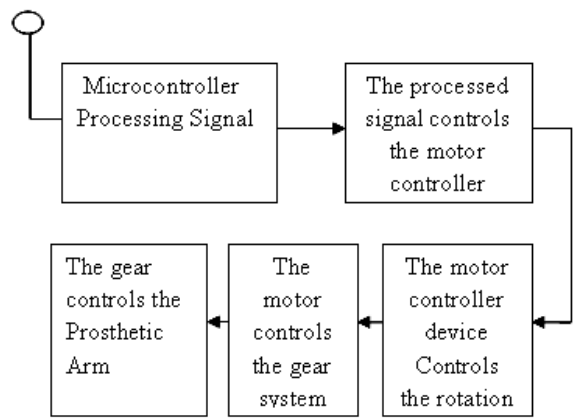

Fig.4: Circuit Block Diagram

### 4. BLOCK DIAGRAM REPRESENTATION OF THE DEVICE

The present prototype is produced by Sensor switch applying the fundamental movement of arm. The signal is sensed by the sensor which is processed by the microcontroller block [2]. Then the signal is passed from microcontroller to the interfacing device. The interfacing device is used for bridging up microcontroller with the motor. The motor control and gear system operates the prosthetic arm.

### 5. THE CONTROLLING CIRCUIT DIAGRAM REPRESENTATION OF THE DEVICE

From the sensor we can detect the signal for arm grasping and stretching. The sensor which has been used is piezo-electric pressure sensor. In the preliminary stage, we are using the soft tact switch for prototype presentation.

The pin2 (RA3), pin3 (RA4) of PIC (16F8FA) catch the signal from the sensor in a digitized form. With the farm-wire programming of micro-controller, we tried to generate a delay for the output port RB0 (pin-6), RB1 (pin-7). In RB0, RB1 the logic 1and 0 is generating for a fixed time with the help of PIC programming. The output of PIC is introduced to L293D, which interfaces the motor controller device. In case of L293D the input pin1 (pin-2) and pin2 (pin-7) received the logic signal in 0 and 1 form a range of values 0.8 V to 6V. The output of the L293D pin1 (pin-3) and pin2 (pin-6) controls the motor for generating arm movement. We are using a DC motor (6V) to develop this design. The motor has a gear facility to obtain the minimum output torque. The controlling arm movement is tractable by the gearing action of motor [7][9].

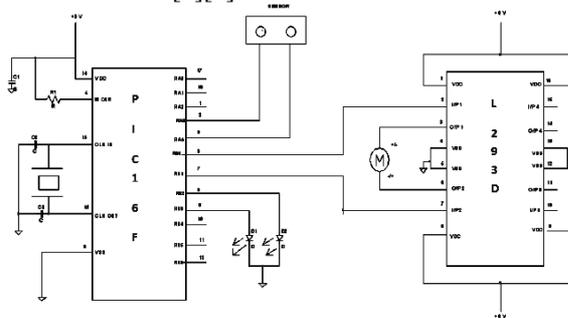

fig 5: Circuit Diagram of motor control unit of prosthetic arm

6. **FLOW CHART OF FARM-WARE PROGRAMMING**

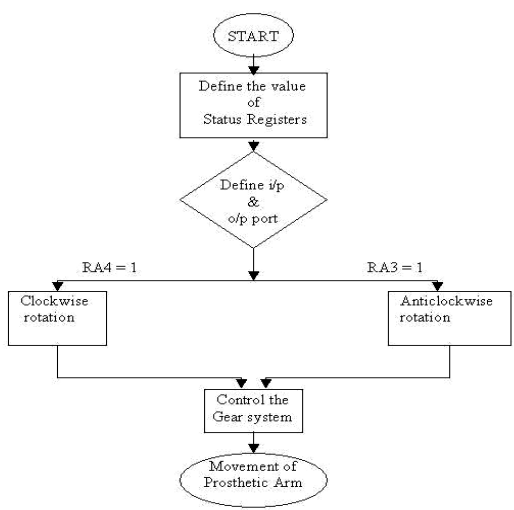

Fig.6.FlowChart

7. **MODELING OF THE DESIGNED PROSTHETIC ARM CONTROL DEVICE**

The formerly described scheme has successfully been implemented by hardware. The main body of the arm consists of hard wood which is cheaper in price and lighter in weight than metallic arm. This mechanical architecture of the arm is like the traditional Otto-Bock arm.

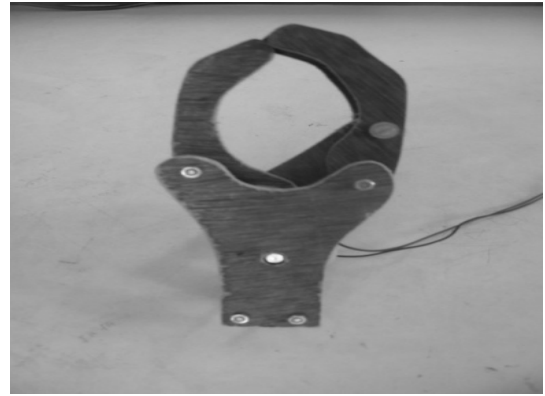

Figure 7A: The Mechanical Body of Prosthetic Arm in Closed Condition

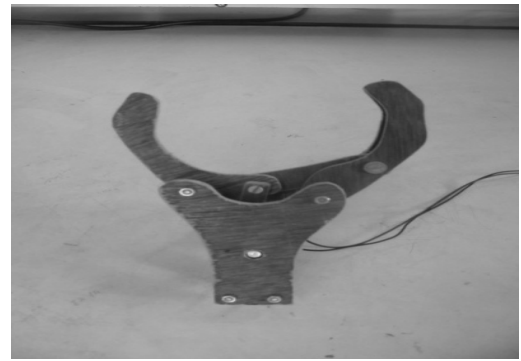

Figure 7B: The Mechanical Body of Prosthetic Arm in Open Condition

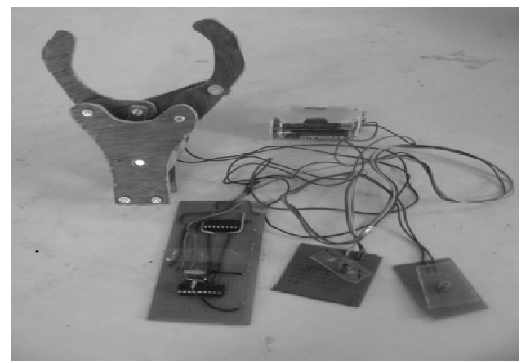

Figure 8: The Prototype Prosthetic Arm with Controlling Accessory and Tact Switch Sensor Circuit

The Fig. 7A and 7B represent the states of the arm in closed gripping condition and open gripping condition respectively. The movement of the prosthetic arm is being controlled by a low power gear motor which produces high torque enhancing the power of griping. The adjoining Fig 8 represents a whole assembly of the prosthetic arm consisting of mechanical wooden body of artificial arm, control circuit, power supply and the sensor switches.

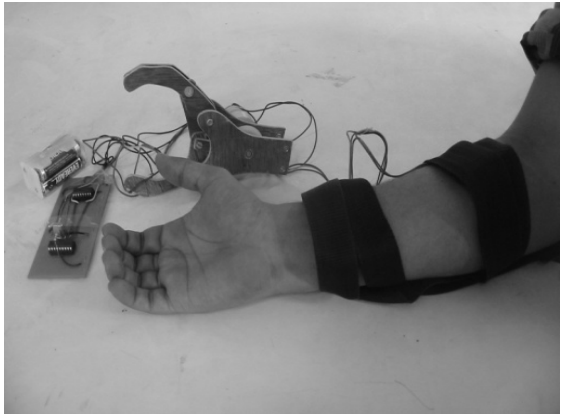

**Figure 9: The Successful Testing of the Prototype Prosthetic Arm with Tact Switch Sensor for Open Condition.**

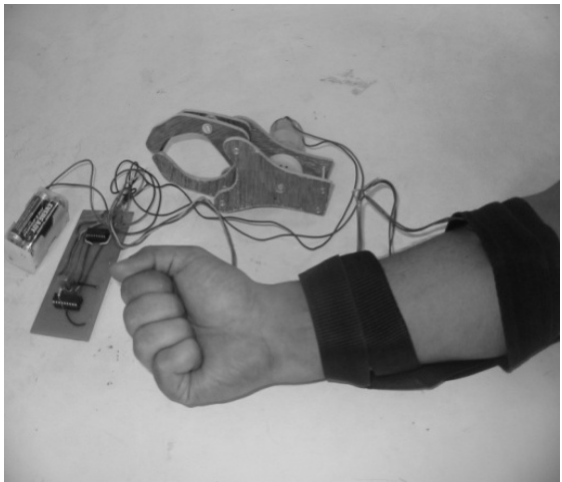

**Figure 10: The Successful Testing of the Prototype Prosthetic Arm with Tact Switch Sensor for Closed Condition.**

The efficacy of this device was successfully authenticated by tying it on a patient hand both in closed gripping condition and open gripping condition. The incorporated tact switch sensed the muscule power and along with the controlling power of the motor, activates the gripping action of the device represented in Fig. 9 and Fig. 10 [1].

## 8. CONCLUSION
This work is based on total hardware implementation of a prosthetic arm model. The transfer function generation and controllability testing are to be considered later on. As the prosthetic control is an emerging phenomenon in today's world, thus this project work will definitely embrace an important role in case of handicapped people. Further developments related to the design and functional accuracies will be the aim for future researchers working in this field.

## 9. ACKNOWLEDGEMENT

The authors acknowledge Dr. P.K Lenka, Professor R&D Lab of NIOH and Mr. A.N. Bishayee Prosthetic Lab, NIOH, Kolkata for their support in recording of EMG signal of patient. The authors owe very much to the R&D Lab of NIOH for lending the facilities of recording through the EMG recorder. We are grateful to acknowledge Dr. S.K. Ray, Professor of ECE Department, Durgapur Institute of Advanced Technology & Management for his encouragement throughout the project work.

## 10. REFERENCE

[1] Dr.Achintya Das & Biswarup Neogi "Prosthetic Control System of Limb using Artificial Intelligence(AI)" International Conference on Systemics, Cybernetics and Informatics ICSCI-2007, pp. 243-247.

[2] Achintya Das, Phd Awarded work on "On Electronics Control of Arterial Pressure of Living Body Employing Bio Feedback Technique"(pp. 103)

[3] The Times of India, Kolkata (pp.13) Wednesday, September 20, 2006.

[4] Biswarup Neogi, Patrali Pradhan & Dr. Achintya Das , National Conference on Recent Trends in Electronics & Communication.(NCRTEC-2008) Mioelectric Signal Detection and Amplification to Develop the Prosthetic Control Model" Vol-1of 2, p.p.244-248.

[5] Careers Eugene J. Moore, Robert M. De Marco, Richard A. Foulds, Tara L. "The Implementation of an EMG Controlled Robotic Arm to Motivate Pre-College Students to Pursue Biomedical Engineering Careers." Pre-College Alvarez Proceedings of 25th Annual International Conference of the IEEE EMBS Cancun, Mexico September 17-21. 2003.

[6] Dudley S. Childress Historical Aspects of Powered Limb Prostheses by Dudley S. Childress, Ph.D. Clinical Prosthetics & Orthotics , 1985, Vol 9, No.1 , pp. 2 – 13.

[7] Dr. Steve C. Hsiung, "The Use of PIC Microcontrollers in Multiple DC Motors Control Applications", Journal of Industrial Technology, Volume 23, Number 3, July 2007 through September 2007. pp. 1-9.

[8] http://www.emgsrus.com/software_emg_graphing.html- EMG GRAPHING.

[9] http://www.alldatasheet.com/datasheet-pdf/pdf/112910/TI/L293D.html.